\documentclass{article}
\usepackage[utf8]{inputenc}
\usepackage[preprint]{corl_2026}
\usepackage{amsmath}
\usepackage{amssymb}
\usepackage{booktabs}
\usepackage{graphicx}
\usepackage{booktabs}
\usepackage{tabularx}
\usepackage{newunicodechar}
\usepackage{placeins}
\usepackage[percent]{overpic}
\newunicodechar{（}{(}
\newunicodechar{）}{)}

\usepackage{xcolor}

\title{ UniDexTok: A Unified Dexterous Hand Tokenizer from Real Data}

\author{
Dong Fang$^{1,2}$ \quad Youjun Wu$^{2,3}$ \quad Yuanxin Zhong$^{2,\ddagger}$ \quad Rui Zhang$^{2,4}$ \quad Yunlong Wang$^{2}$ 
\\ \textbf{Xiaosong Jia}$^{1,\dagger}$ 
\quad \textbf{Yu-Gang Jiang}$^{1,\dagger}$ \\
\\
$^{1}$Fudan University \quad $^{2}$Rimbot \\
$^{3}$Hefei University of Technology \quad $^{4}$Beijing University of Posts and Telecommunications 
}

\begin{document}
\maketitle

\begingroup
\renewcommand{\thefootnote}{\fnsymbol{footnote}}
\footnotetext[1]{This work was done when Dong Fang, Youjun Wu and Rui Zhang were research interns at Rimbot.}
\footnotetext[2]{Corresponding authors.}
\footnotetext[3]{Project leader.}
\endgroup

\begin{abstract}
 Dexterous hands are essential for fine \textbf{}grained manipulation, but their hardware designs vary substantially across embodiments. Differences in kinematics, joint definitions, and degrees of freedom make it difficult to define a shared state representation, compared to gripper. As a result, dexterous hand data remains fragmented and difficult to use for joint training. In this work, we propose the Unified Dexterous Hand Model (UDHM), which maps human and robot hand states into a shared 22-DoF semantic interface. Based on UDHM, we introduce UniDexTok, a retarget-free state tokenizer that learns embodiment conditioned discrete tokens from standardized real joint states. UniDexTok provides a unified representation for heterogeneous dexterous hands without relying on retargeting or simulation data. Compared with the recent baseline UniHM, UniDexTok reduces MPJAE from 15.63 degrees to 0.16 degrees and MPJPE from 18.51 mm to 0.18 mm, corresponding to error reductions of 98.98\% and 99.03\%, respectively. These results improve reconstruction from centimeter-scale to sub-millimeter accuracy. Experiments further show that data from other embodiments improves target embodiment reconstruction accuracy, demonstrating the benefit of cross embodiment tokenization. Also, UniDexTok shows an strong zero-shot and few-shot reconstruction ability, when new dexterous hands are introduced.
\end{abstract}


\begin{figure}[htbp]
    \centering
    \includegraphics[width=0.9\linewidth]{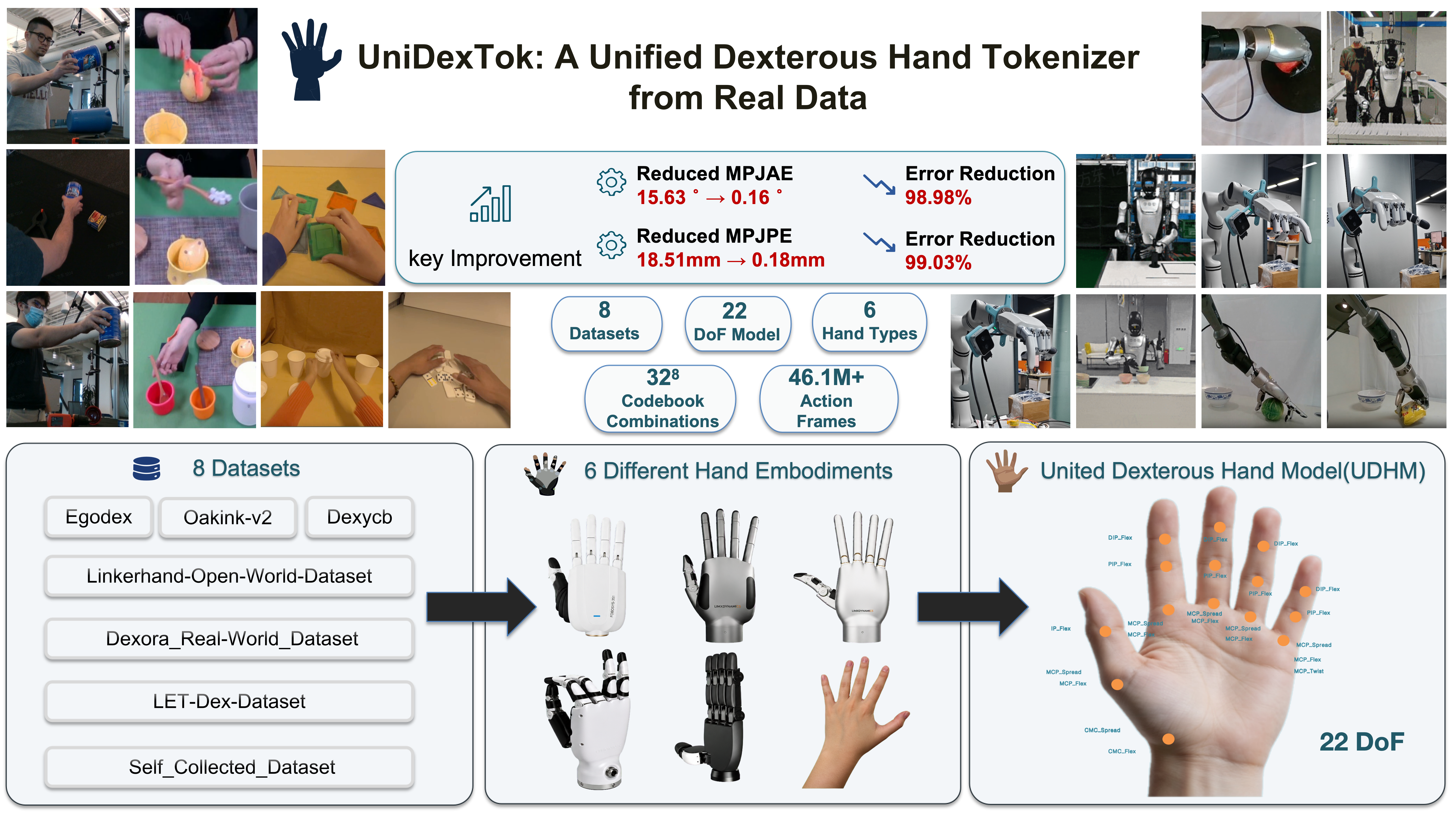}
    \caption{Overview of UniDexTok.}
    \label{fig:overview}
\end{figure}

\section{Introduction}

Embodied intelligence~\citep{long2025embodied,gupta2021embodied} has recently attracted increasing attention, yet current vision-language-action models~\citep{brohan2023rt2,kim2024openvla,grover2025genvla,xlvla2026,wen2025grdextertechnicalreport} and world-action models~\citep{cen2025worldvla,ye2026gigaworld,kim2025dwm,kim2026cosmospolicy,goswami2026dexwm} still struggle with tasks that require fine-grained manipulation. While recent work has explored efficient action tokenization for VLA models~\citep{pertsch2025fast}, these approaches focus on action sequences rather than robot state representations. To meet the growing demand for human-like dexterous manipulation, the community is increasingly moving from parallel grippers toward dexterous hands~\citep{billard2019dexterous,shaw2023leaphand,huang2025humanlike}. However, progress in dexterous-hand learning is hindered by the large variation in hardware configurations, joint definitions, control conventions, and degrees of freedom across different hand embodiments. These inconsistencies make it difficult to share data, train models jointly, or build scalable dexterous manipulation systems.

Motivated by the observation that modern dexterous-hand designs are increasingly converging toward human-hand-like kinematic structures, we propose a Unified Dexterous Hand Model (UDHM, Figure~\ref{fig:overview}), a standardized hand representation that covers a broad range of dexterous-hand embodiments, including the human hand. UDHM treats the human hand as an additional dexterous embodiment and maps both human and robot-hand states into a shared active-joint interface. This enables heterogeneous dexterous-hand datasets to be used jointly while preserving embodiment-specific kinematic information. Our experiments further show that incorporating human-hand data~\citep{yang2022oakink,hoque2026egodex,chao2021dexycb} significantly improves model performance.

\begin{figure}[t]
    \centering
    \includegraphics[width=1.0\linewidth]{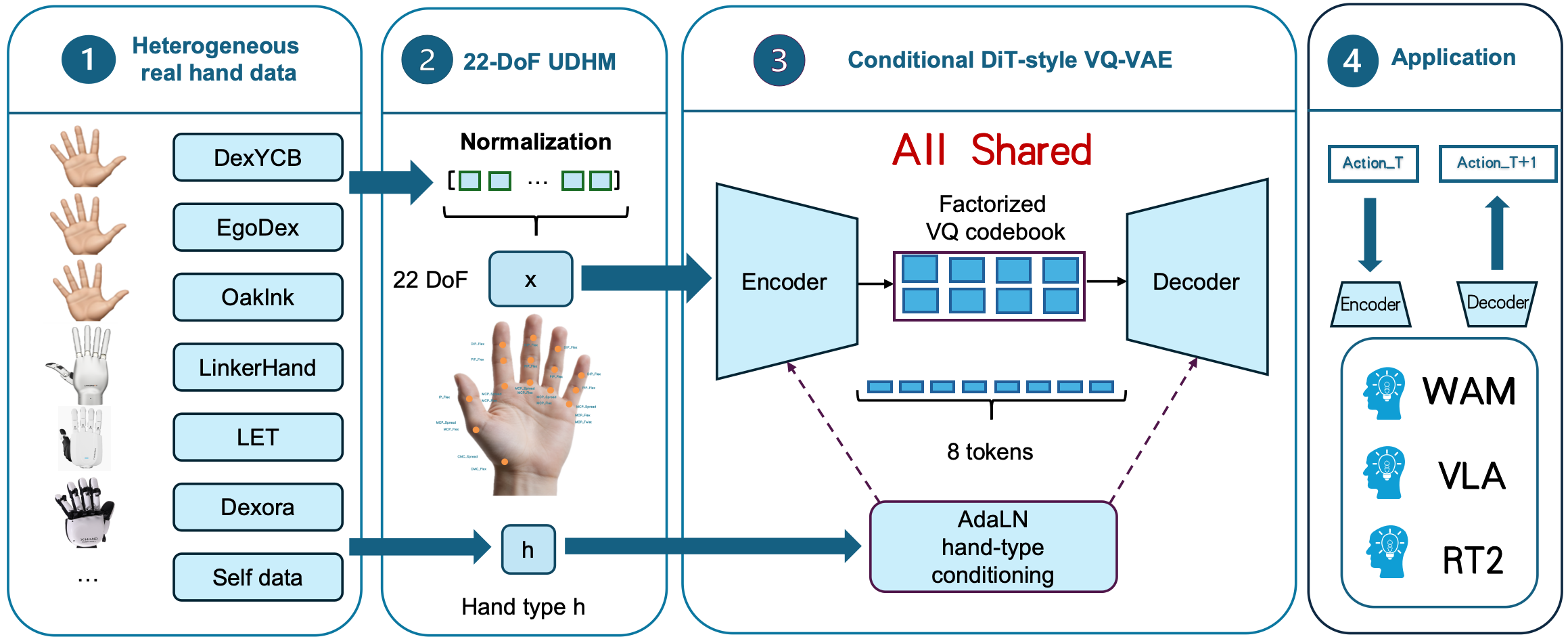}
    \caption{UniDexTok architecture. A conditional transformer encoder maps standardized hand states to latent tokens, factorized vector  squantization discretizes them, and a conditional decoder reconstructs the continuous state.}
    \label{fig:architecture}
\end{figure}

Building on UDHM, we introduce UniDexTok, a unified tokenizer for dexterous-hand states across different hand morphologies. Unlike UniHM~\citep{zhang2026unihm} and UniDex~\citep{zhang2026unidex}, which train separate hand-specific tokenizers for each robot hand, UniDexTok shares a single encoder, codebook, and decoder across all embodiments, as shown in Figure~\ref{fig:architecture}. With UDHM, UniDexTok shares not only the codebook but also the encoder and decoder across embodiments. The shared encoder maps heterogeneous hand states into a common latent coordinate system before quantization, and the shared decoder gives discrete tokens a consistent cross-hand interpretation. This design encourages transferable structure across different hand morphologies and avoids learning isolated hand-specific latent spaces. As a result, an unseen hand can be projected into the existing token space without training a new tokenizer from scratch, enabling zero-shot transfer~\citep{zha2026lap} and efficient few-shot adaptation~\citep{li2025controlvla} with limited target-domain data. Reconstruction experiments show that this unified tokenizer substantially outperforms UniHM, reducing both joint-angle and Cartesian reconstruction errors across multiple dexterous hands.

Building on UDHM and UniDexTok, we introduce a retargeting-free dexterous-hand state learning pipeline that, to our knowledge, is the first to train on real robot-hand data spanning multiple dexterous-hand datasets~\citep{hoque2026egodex,let2026,yang2022oakink,dexora2026,linkerhand2026,chao2021dexycb} by standardizing them into a unified representation. The pipeline first standardizes heterogeneous human and robot hand states with UDHM, and then learns shared discrete state tokens with UniDexTok. This standardization step is essential because the main bottleneck of real dexterous-hand data is not only data quantity, but also data usability: existing datasets differ in joint order, control order, dimensionality, and units, including radians, degrees, and discretized bins. In some cases, even datasets collected from the same hand use different recording and control conventions. We unify these heterogeneous datasets under a common standard and learn directly from standardized real hand states, rather than obtaining robot-hand data by retargeting~\citep{maniptrans2025,xin2025retargeting} MANO~\citep{romero2017mano} trajectories, which can introduce geometric mismatch, or by relying on simulation~\citep{wang2023dexgraspnet}, which suffers from the sim-to-real gap~\citep{pan2026beyond,zeng2026dexsim2real,hsieh2025dexscrew}. To our knowledge, this is the first robot state tokenizer trained on real cross-embodiment dexterous-hand data.

Recent advances in diffusion policy~\citep{zhu2025scaledp,song2025diffusion} have shown promising results for learning visuomotor robot control. However, these methods primarily focus on end-effector control and lack the fine-grained tactile sensing~\citep{guo2025tactile} capabilities essential for dexterous manipulation. Our unified tokenizer provides a foundation for future integration with these policy learning frameworks, enabling more sophisticated dexterous manipulation skills.

The contributions of this paper are as follows:
\begin{itemize}
    \item We propose UDHM, a unified hand model that maps human hands and diverse dexterous robot hands into a shared joint space while preserving embodiment-specific kinematic information.

    \item We introduce UniDexTok, a unified cross-embodiment tokenizer that shares a single encoder, codebook, and decoder across hand embodiments, forming a common discrete token space for zero-shot transfer and efficient few-shot adaptation to unseen hands.

    \item We present the first retargeting-free pipeline training tokenizer directly on standardized real dexterous-hand data spanning multiple datasets and hand embodiments.
\end{itemize}

\section{Related Work}

\paragraph{Retargeting-based dexterous data generation.}
Many recent methods address the scarcity of dexterous-hand data by converting human demonstrations into robot-hand trajectories.
UniHM uses MANO hand data as the source and retargets the same human motion to several robot hands, such as Shadow, Allegro, and LEAP~\citep{shaw2023leaphand}, so that different hands receive approximately paired action sequences~\citep{zhang2026unihm}.
UniDex builds a robot-centric corpus from egocentric RGB-D hand-object interaction datasets such as H2O, HOI4D, HOT3D, and TACO by retargeting human motions into executable dexterous-hand trajectories~\citep{zhang2026unidex}.
Its pipeline aligns fingertips through inverse kinematics, introduces a 6-DoF dummy base offset for palm adjustment, and handles mimic-joint constraints for hands such as Inspire and Oymotion.
Latent Action Diffusion learns paired actions across different end effectors through retargeted human poses~\citep{bauer2026latent}.
CrossDex uses MANO eigengrasps as a unified action space and retargets the resulting hand pose into the joint positions of each robot hand~\citep{yuan2024crossdex}.
These approaches are effective for producing executable demonstrations, but the retargeting step necessarily changes the native state and can introduce geometric mismatch.

\paragraph{Simulation-based dexterous data generation.}
Another line of work uses simulation to synthesize dexterous grasping data at scale.
DexGraspNet generates ShadowHand grasps for many 3D objects using a differentiable force-closure estimator and validates grasp stability in Isaac Gym~\citep{wang2023dexgraspnet}.
DexGraspNet 2.0 extends this idea to large-scale cluttered scenes and synthetic grasp candidates~\citep{zhang2024dexgraspnet2}.
CrossDex also trains cross-embodiment grasping policies in Isaac Gym~\citep{yuan2024crossdex}.
Simulation offers coverage that real datasets do not yet provide, but it introduces a sim-to-real gap and does not solve the problem of standardizing real robot-hand logs that already exist in the community.
UniDexTok is complementary to these methods: it focuses on a retarget-free state representation for real heterogeneous hand data.

\section{Method}

UniDexTok has three components: a unified hand-state representation, a conditional tokenizer architecture, and a factorized vector-quantization codebook.
The representation defines a shared state space across embodiments, the tokenizer maps continuous states to discrete tokens and reconstructs them back to continuous states, and the codebook controls the capacity of the discrete vocabulary.

\subsection{Unified Dexterous Hand Model}

UDHM parameterizes hand pose with anatomically motivated kinematic constraints so that MANO-format 21-joint human-hand keypoints can be represented by a low-dimensional and interpretable degree-of-freedom vector.
Given one frame of hand joint coordinates, UDHM uses a straight, adducted hand as the zero pose and assumes that the palm is rigid.
Under this assumption, the thumb CMC joint and the four non-thumb MCP joints keep fixed offsets relative to the wrist.

\begin{figure}[t]
    \centering
    \includegraphics[width=0.75\linewidth]{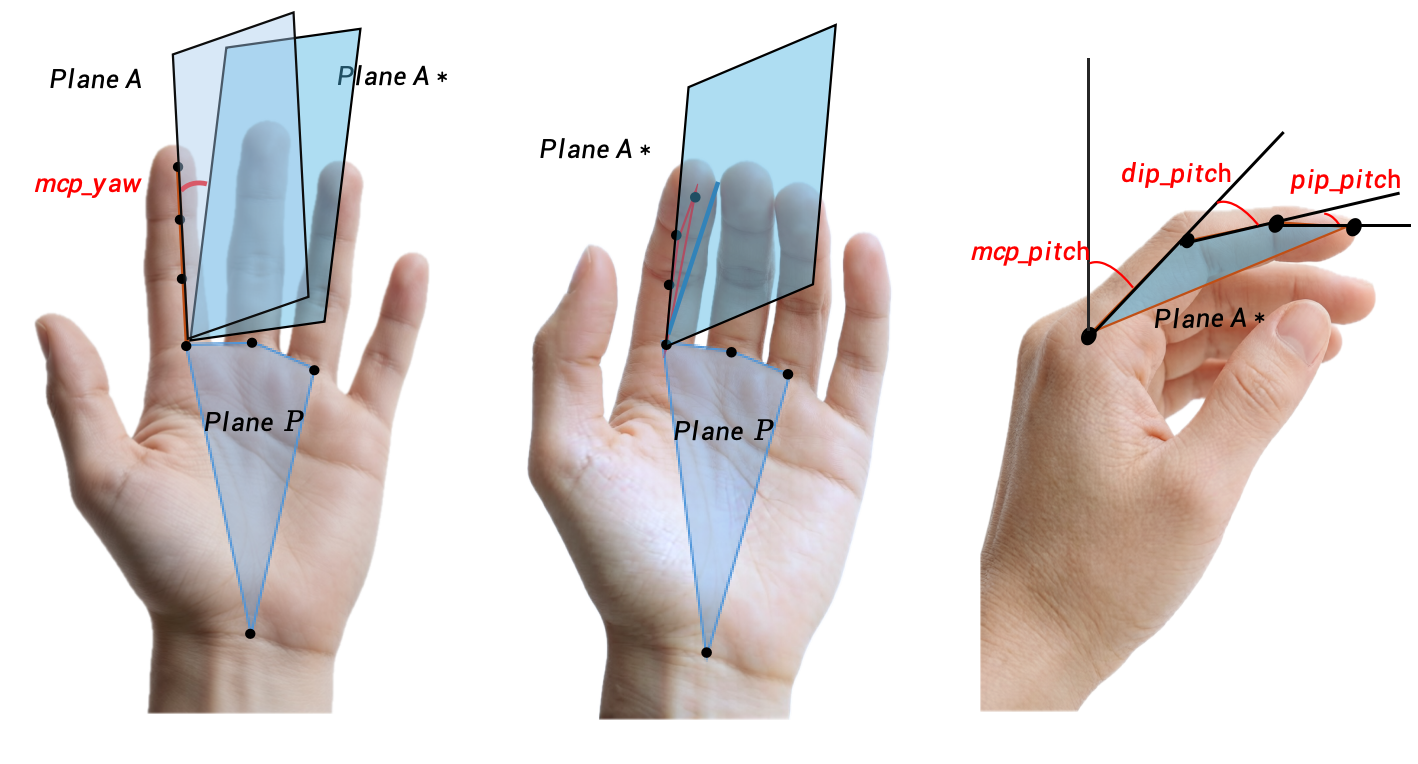}
    \caption{UDHM kinematic parameterization. The model fits a palm plane, defines local motion axes, and reconstructs each finger chain with analytically defined forward kinematics.}
    \label{fig:UDHM_fk}
\end{figure}

As illustrated in Figure~\ref{fig:UDHM_fk}, the palm plane $P$ is fitted by the Wrist, Index MCP, Middle MCP, and Ring MCP keypoints, and the plane normal $n_p$ is used as the palm normal.
For the index, middle, and ring fingers, UDHM uses four degrees of freedom: MCP abduction/adduction, MCP flexion/extension, PIP flexion/extension, and DIP flexion/extension.
The MCP abduction/adduction rotates around the palm normal $n_p$, while the flexion motions rotate around a local lateral axis $e_{\mathrm{lat}}$.
This construction keeps each MCP--PIP--DIP--tip chain in a local motion plane that is perpendicular to the palm plane.
The pinky adds one twist degree of freedom around the metacarpal direction to compensate for its larger lateral pose variation.
The thumb uses five degrees of freedom: CMC flexion, CMC spread, MCP flexion, MCP abduction, and IP flexion, with the thumb motion plane defined dynamically from the CMC--MCP direction.
These choices give 22 active coordinates in total.

UDHM is a compact hand representation with forward and inverse maps to MANO-format keypoints.
In forward kinematics, the wrist joint is fixed first.
The MCP, PIP, DIP, and fingertip positions are then computed sequentially with bone lengths extracted from the input joints and Rodrigues axis-angle rotations.

For inverse mapping, the wrist position is directly taken from the input keypoint:
\begin{equation}
    p_w = J_0 .
\end{equation}
UDHM then extracts palm offsets and bone lengths from the input frame and refines the joint angles by solving a nonlinear least-squares problem, so that the forward-kinematics reconstruction matches the target joints.

In a test on a 100-frame MANO keypoint sequence, UDHM obtains an negligible reconstruction error.
The remaining error mainly comes from the rigid-palm assumption and from the coplanar or perpendicular local motion-plane constraints used for the index, middle, and ring fingers.

The same representation is used by the multi-source dataset standardization pipeline.
Joint records from different sources are mapped to the shared 22-dimensional order and paired with a hand-type label $h$.
Thus, the tokenizer receives real hand states in a unified representation rather than states produced by MANO fitting across hands or by retargeting between different robot embodiments.
All raw joint angles are represented in radians and normalized by a fixed scale of \(\pi\).
This fixed normalization avoids source-specific statistics that would make tokens depend on a particular dataset split, while reconstructed metric, MPJAE remains directly interpretable after denormalization.
An example showing the necessity of normalization is that, LET and Linkhand datasets, LET-Dex-Dataset and LinkerHand-Open-World-Dataset logs encode some joints as values in a 0--255 bin range, so they are first converted into the same radian-based joint-angle format.

\subsection{Conditional State Tokenizer}

Let $x \in \mathbb{R}^{D}$ be a single hand state and let $h$ denote a discrete hand-type or data-source label.
In this work, $D=22$, matching the active UDHM interface.
The tokenizer learns a shared encoder $E$, quantizer $Q$, and decoder $D$.
It reconstructs each state after fixed angle normalization:
\begin{equation}
    \hat{x} = \pi D(Q(E(\tilde{x}, h_{embed})), h_{embed}),
    \quad \tilde{x}=x/\pi .
\end{equation}

Although the hand state is low-dimensional, the joint motion required for a task depends on the hand morphology. The same functional action can require different flexion, spread, and opposition patterns for hands with different finger lengths or palm geometry, creating correlations between action coordinates and finger-size structure. We therefore use a transformer tokenizer inspired by DiT-style blocks, adapted from image latents to kinematic states, to model these morphology-dependent correlations.
The encoder first projects the normalized state into $N$ latent tokens:
\begin{equation}
    z_0 = \mathrm{reshape}(W_e \tilde{x} + b_e) + p,
    \quad z_0 \in \mathbb{R}^{N \times C},
\end{equation}
where $p$ is a learned positional embedding.
The reported model uses $N=8$ and $C=512$.
The token sequence is passed through transformer blocks with self-attention, MLP layers, zero-initialized residual gates, and adaptive layer normalization.
When a hand-type label is available, the embedding $c_h$ modulates the layer-normalization scale and shift:
\begin{equation}
    \mathrm{AdaLN}(z,h_{embed}) =
    \gamma(c_h) \odot \mathrm{LN}(z) + \beta(c_h).
\end{equation}
This conditioning lets the shared token space model functional hand states while the encoder and decoder still account for hand-specific kinematic conventions.

The decoder mirrors the encoder.
Quantized tokens are projected back to the transformer width, positional embeddings are added, and conditional transformer blocks reconstruct the normalized state.
The final output head flattens token features and predicts $\hat{x}\in\mathbb{R}^{22}$ after denormalization.
The reconstruction loss combines mean squared error with a SmoothL1 auxiliary term in normalized angle space:
\begin{equation}
    \mathcal{L}_{rec}
    =
    \mathrm{MSE}(\tilde{x}, \hat{\tilde{x}}).
\end{equation}

\subsection{Factorized Codebook Design}

A single 256-entry codebook can represent only 256 discrete states for each quantized vector.
UniDexTok factorizes the vocabulary over channel groups.
The encoder output is projected from 512 to 256 dimensions and divided into $K=8$ channel groups.
Each group is quantized by a 32-entry sub-codebook.
As a result, one token can express $32^8$ code combinations while using only $32 \times 8=256$ learned code vectors.
With $N=8$ token positions, the representation composes this factorized structure over the whole hand.

For group $k$, the quantizer selects the nearest normalized code vector by cosine similarity:
\begin{equation}
    i_{n,k}
    =
    \arg\max_j
    \left\langle
    \frac{u_{n,k}}{\|u_{n,k}\|_2},
    \frac{e_{k,j}}{\|e_{k,j}\|_2}
    \right\rangle ,
\end{equation}
where $u_{n,k}$ is the projected feature of token $n$ in group $k$.
The quantized vector is obtained by concatenating the selected sub-code vectors.
Training uses the straight-through estimator and the standard VQ commitment loss~\citep{oord2018vqvae}:
\begin{equation}
    \mathcal{L}_{vq}
    =
    \beta \| \mathrm{sg}[q] - u \|_2^2
    +
    \| q - \mathrm{sg}[u] \|_2^2 .
\end{equation}
We set $\beta=0.25$ in all reported experiments.
The implementation supports entropy regularization, but the final model disables it because reconstruction quality is the primary objective.
The total loss is
\begin{equation}
    \mathcal{L}
    =
    \mathcal{L}_{rec}
    +
    \mathcal{L}_{vq}.
\end{equation}
Checkpoints are selected by raw joint-angle MAE in degrees, computed after multiplying normalized predictions by $\pi$.

\section{Experiments}

\subsection{Datasets}
The dataset contains three groups. All groups are split 80\%/20\% into train and test.

The first is coarse-grained human hand-object interaction data (DexYCB~\citep{chao2021dexycb}, OakInk-v2~\citep{yang2022oakink}, EgoDex~\citep{hoque2026egodex}): rather than retargeting it, we reduce the MANO-style 45-DoF representation to the 22 active UDHM coordinates and train directly on the standardized human-hand.

The second group consists of real-world public dexterous-hand datasets (LET~\citep{let2026}, Dexora~\citep{dexora2026}, LinkerHand~\citep{linkerhand2026}). Although LET, Dexora, and LinkerHand differ in joint ordering, dimensionality, and units, their data can be converted into a unified format through standardization. We convert all joint values to radians, insert each available DoF into the semantically corresponding UDHM coordinate, and zero-pad the missing coordinates. We refer to this mapping as semantic insertion and compare it with an append-and-pad alternative in the ablation study, as shown in Table~\ref{tab:semantic_insertion}.

\subsection{Metrics}

We evaluate hand-state quality in both joint-angle space and Cartesian space.
Mean Per-Joint Angle Error (MPJAE) measures the average absolute angle joint difference between the reconstructed and the target hand.
Mean Per-Joint Position Error (MPJPE) measures the average Euclidean distance between reconstructed and target hand joints.
Besides, we use 'FK error' to denote the fingertip-position errors

\subsection{Main Results}

We compare UniDexTok with UniHM, the closest recent cross-hand tokenizer baseline. For a fair comparison, we follow the data construction protocol described in the UniHM paper. Specifically, we retarget DexYCB hand states to multiple dexterous hands, including LinkerHand L6, LinkerHand L10, LinkerHand L20, and Robotera XHand1, to obtain robot-hand training data. We then train a separate encoder and decoder for each hand, following the UniHM setting. Because UniHM is trained with retargeted data derived from sources such as DexYCB, we adopt two evaluation protocols to ensure a fair comparison.
The first protocol evaluates on DexYCB-derived retargeted test states for both UniHM and UniDexTok.
Since the public UniHM release does not specify the exact 80/20 split used during training, we conservatively treat all available UniHM data as test data; this favors UniHM because it may includes its training sequences. Moreover, our model has never used data obtained from DexYCB retargeting, which makes this setting equivalent to zero-shot for our model.
The second protocol evaluates UniHM and UniDexTok on the 20\% held-out split from all standardized datasets.

\begin{table}[!htbp]
\caption{Reconstruction comparison on various hands on both our datasets (no retargeting) and the retargeted DexYCB dataset.}
\label{tab:combined_results}
\centering
\small
\resizebox{0.8\textwidth}{!}{
\begin{tabular}{lcccccccc}
\toprule
& \multicolumn{4}{c}{(a) on our datasets} & \multicolumn{4}{c}{(b) on the retargeted DexYCB dataset} \\[2pt]
& \multicolumn{2}{c}{MPJAE (deg) $\downarrow$} & \multicolumn{2}{c} {MPJPE (mm) $\downarrow$} & \multicolumn{2}{c}{MPJAE (deg) $\downarrow$} & \multicolumn{2}{c}{MPJPE (mm) $\downarrow$} \\
\cmidrule(lr){2-3}\cmidrule(lr){4-5}\cmidrule(lr){6-7}\cmidrule(lr){8-9}
\textbf{Hand} & UniHM & Ours & UniHM & Ours & UniHM & Ours & UniHM & Ours \\
\midrule
LinkerHand L6 & 16.81 & 0.15 & 13.79 & 0.15 & {\color[HTML]{333333} 1.22} & 0.42 & 1.13 & 1.09 \\
LinkerHand L10 & 17.95 & 0.15 & 12.69 & 0.13 & 5.80 & 3.05 & 4.72 & 4.86 \\
LinkerHand L20 & 16.92 & 1.08 & 14.40 & 0.91 & 9.25 & 7.04 & 6.75 & 6.58 \\
Robotera XHand1 & 15.10 & 0.15 & 20.31 & 0.18 & 1.34 & 0.79 & 1.47 & 1.35 \\
\midrule
All & 15.63 & 0.16 & 18.51 & 0.18 & 4.40 & 2.83 & 3.52 & 3.47 \\
\bottomrule
\end{tabular}
}
\end{table}

Table~\ref{tab:combined_results} reports the core reconstruction comparison under the two evaluation protocols. On our datasets, UniDexTok reduces the average MPJAE from 15.63 degrees to 0.16 degrees and reduces the average MPJPE from 18.51 mm to 0.18 mm.  The improvement is consistent across LinkerHand L6, LinkerHand L10, LinkerHand L20, and Robotera XHand1, which suggests that UniDexTok does not only fit one specific hand type. On the retargeted DexYCB dataset, UniDexTok also obtains lower average MPJAE than UniHM, although the MPJPE gap is smaller. This result is reasonable because this protocol is closer to the retargeted data distribution used by UniHM. More importantly, the result supports the main design choice of UniDexTok. The shared encoder learns a common representation space across different hand embodiments. This unified encoder is the prerequisite for later zero-shot and few-shot transfer, because a new hand can be mapped into the same semantic state space instead of being learned as an isolated embodiment. Therefore, Table~\ref{tab:combined_results} does not only demonstrate reconstruction accuracy. It also shows that the unified encoder provides the representational foundation that makes cross-embodiment generalization possible.

\textbf{This result supports the central hypothesis of the paper: for state representation learning, preserving real hand states in a shared semantic space is more accurate than first retargeting those states into another embodiment.}

\begin{table}[!htbp]
\caption{Zero-shot and few-shot reconstruction errors with Inspire hand. Joint-angle errors are reported in degrees, and fingertip-position errors are reported as FK error.}
\label{tab:zero_few_shot}
\centering
\footnotesize
\setlength{\tabcolsep}{4pt}
\renewcommand{\arraystretch}{0.95}

\resizebox{0.95\textwidth}{!}{
\begin{tabular}{lccc@{\hspace{14pt}}lccc}
    \toprule
    \multicolumn{4}{c}{MPJAE (deg) $\downarrow$} & \multicolumn{4}{c}{FK Error (mm) $\downarrow$} \\
    \cmidrule(lr){1-4}\cmidrule(lr){5-8}
    \textbf{Joint} & \textbf{Zero-shot} & \textbf{Few-shot} & \textbf{Error Reduction}
    & \textbf{Finger} & \textbf{Zero-shot} & \textbf{Few-shot} & \textbf{Error Reduction} \\
    \midrule
    Thumb CMC pitch  & 4.91 & 1.85 & 62.4\% & Thumb  & 11.88 & 4.93 & 58.5\% \\
    Thumb CMC yaw    & 5.44 & 1.69 & 69.0\% & Index  & 16.05 & 3.40 & 78.8\% \\
    Index MCP pitch  & 7.85 & 1.68 & 78.6\% & Middle & 13.43 & 3.73 & 72.2\% \\
    Middle MCP pitch & 6.29 & 1.76 & 72.1\% & Ring   & 11.30 & 4.15 & 63.3\% \\
    Ring MCP pitch   & 5.58 & 2.08 & 62.8\% & Pinky  & 7.73  & 2.65 & 65.7\% \\
    Pinky MCP pitch  & 4.14 & 1.42 & 65.7\% &        &       &      &        \\
    \bottomrule
\end{tabular}}

\end{table}

We further evaluate the generalization capability of UniDexTok to a new hand embodiment not seen during training. Using the Inspire hand RH56E2, which features six active joints with a kinematic structure distinct from all hands in the training set, we test both zero-shot transfer and few-shot fine-tuning. The unified encoder, trained on diverse multi-embodiment data, establishes a shared representation space that serves as the prerequisite for effective zero-shot and few-shot generalization. For few-shot adaptation, we fine-tune on only 4,528 frames (6.2\% of the full Inspire dataset) for 2 epochs, without any architectural modification. As shown in Table~\ref{tab:zero_few_shot}, in the zero-shot setting UniDexTok achieves per-joint angle errors between 4.14 and 7.85 deg, with the largest error on Index MCP pitch. Fingertip FK errors range from 7.73 to 16.05 mm. After few-shot adaptation, the errors of all six active joints decrease substantially. Thumb CMC pitch and yaw decrease to 1.85 and 1.69 deg, corresponding to error reductions of 62.4\% and 69.0\%, respectively. The four non-thumb MCP pitch errors are all below 2.08 deg, with Index MCP pitch showing the largest relative error reduction of 78.6\%. Fingertip FK errors are also reduced by 58.5--78.8\%, with the Index fingertip error decreasing from 16.05 to 3.40 mm. These results indicate that UniDexTok can transfer to an unseen hand embodiment and can be adapted with a small fraction of target-domain data, substantially reducing the hand-to-hand domain gap.

\textbf{These results suggest that the unified tokenizer, consisting of a shared encoder, codebook, and decoder, is the key factor enabling both zero-shot transfer and few-shot adaptation. By learning a common discrete state space across hand embodiments, UniDexTok allows an unseen hand to be mapped into the same token representation and then adapted with only a small amount of target-domain data.}

\subsection{Visualization}

\begin{figure*}[!htbp]
    \centering
    \setlength{\parskip}{0pt}

    \begin{overpic}[width=1.0\linewidth]{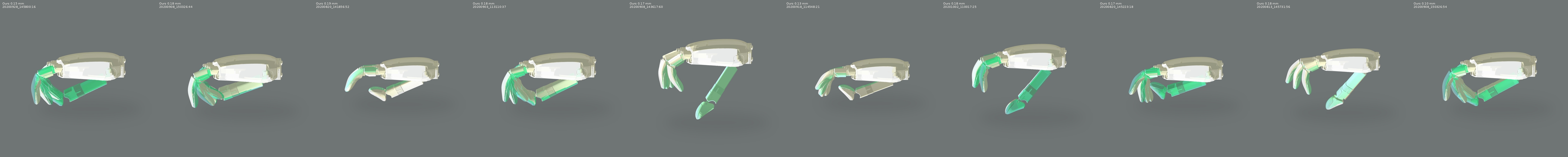}
        \put(1,7.5){\small\textbf{(a)}}
    \end{overpic}\\[-0.2em]
    \includegraphics[width=1.0\linewidth]{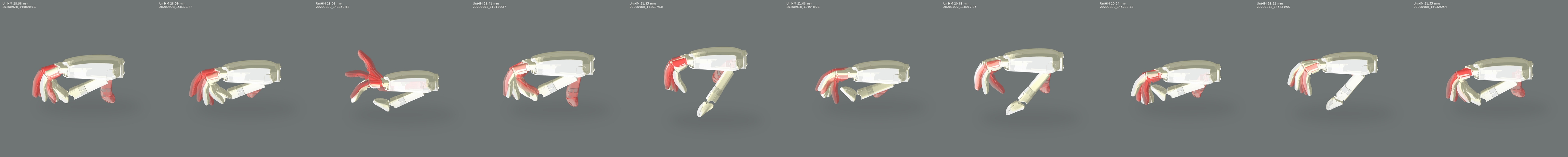}\\[-0.2em]
    \begin{overpic}[width=1.0\linewidth]{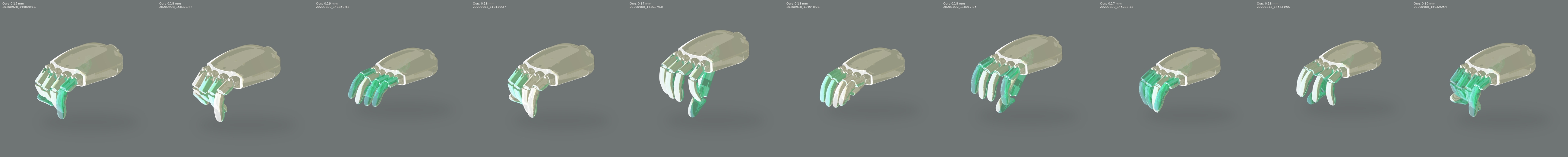}
        \put(1,7.5){\small\textbf{(b)}}
    \end{overpic}\\[-0.2em]
    \includegraphics[width=1.0\linewidth]{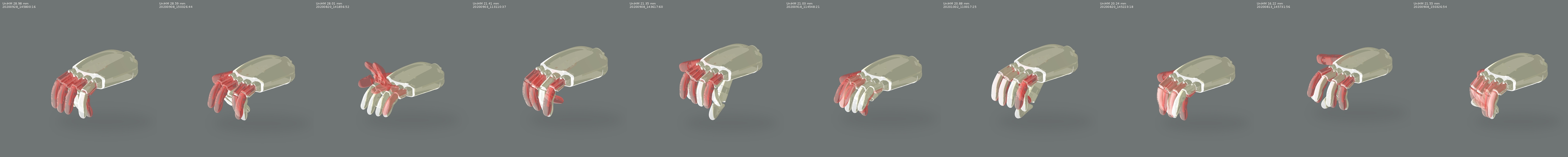}

    \caption{
    Visual comparison of reconstruction results across different views. (a) Side view of UniDexTok and UniHM reconstruction. 
    (b) Oblique view of UniDexTok and UniHM reconstruction. White hands denote ground-truth states, green hands denote UniDexTok reconstructions, and red hands denote UniHM reconstructions.
    }
    \label{fig:visualize_reconstruction}
\end{figure*}

To provide a qualitative comparison between UniDexTok and UniHM, as shown in Figure~\ref{fig:visualize_reconstruction}, we randomly select 10 hand states from different hand embodiments and visualize the reconstruction results in the SAPIEN simulator. The white hand denotes the ground-truth state before reconstruction, the green hand denotes the UniDexTok reconstruction, and the red hand denotes the UniHM reconstruction. These visual comparisons show that UniDexTok reconstructs the target hand states more accurately than UniHM.

\subsection{Ablation Study}

\begin{table}[!htbp]
    \caption{Reconstruction comparison with and without human-hand data.}
    \label{tab:unidextok_hand}
    \centering
    \resizebox{0.9\textwidth}{!}{
    \begin{tabular}{lcccc}
        \toprule
        & \multicolumn{2}{c}{ MPJAE (deg) $\downarrow$} & \multicolumn{2}{c}{MPJPE (mm) $\downarrow$} \\
        \cmidrule(lr){2-3}\cmidrule(lr){4-5}
        \textbf{} & UniDexTok & UniDexTok w/o human-hand & UniDexTok & UniDexTok w/o human-hand \\
        \midrule
        LinkerHand L6 & {\color[HTML]{333333} 0.15} & 0.27 & 0.15 & 0.27 \\
        LinkerHand L10 & 0.15 & 0.35 & 0.13 & 0.32 \\
        LinkerHand L20 & 1.08 & 1.75 & 0.91 & 1.57 \\
        Robotera XHand1 & 0.15 & 0.38 & 0.18 & 0.46 \\
        \midrule
        All & 0.16 & 0.37 & 0.18 & 0.43 \\
        \bottomrule
    \end{tabular}}
\end{table}

We evaluate two design choices in UDHM standardization.
The first ablation studies whether human-hand data should be included as an additional dexterous embodiment. As shown in Table~\ref{tab:unidextok_hand}, adding UDHM-processed human hand-object data improves model learning because it expands pose coverage while keeping all states in the same 22-dimensional semantic interface.
\textbf{This supports the view that the human hand should not only be treated as a retargeting source, but also as a valid training embodiment for state-token learning.}

\begin{table}[!htbp]
    \caption{Reconstruction comparison with and without Semantic insertion.}
    \label{tab:semantic_insertion}
    \centering
    \resizebox{0.9\textwidth}{!}{
    \begin{tabular}{lccc}
        \toprule
        Method & MPJAE (deg) $\downarrow$ & MPJPE (mm) $\downarrow$ & FK Error (mm) $\downarrow$ \\
        \midrule
        UniDexTok & 0.24 & 0.25 & 0.53 \\
        \begin{tabular}[c]{@{}l@{}}UniDexTok w/o semantic-insertion\end{tabular}
        & 0.53 & 0.57 & 1.20 \\
        \bottomrule
    \end{tabular}}
\end{table}

The second ablation compares two mappings for robot-hand dimensions.
As shown in Table~\ref{tab:semantic_insertion}, in the append-and-pad variant, all available robot-hand coordinates are placed at the front of the vector and the remaining dimensions are padded with zero.
In the semantic-insertion variant used by UniDexTok, each available robot-hand coordinate is inserted into its corresponding UDHM dimension, and missing joints are padded with zero.
The semantic-insertion variant performs better, indicating that the model is not merely learning hand-specific numerical compression.
Instead, it benefits from the joint-level semantic alignment provided by the unified hand-state space.



\FloatBarrier

\subsection{Representation Quality}

To evaluate the semantic discriminability of the learned representations, we construct a 13-class gesture classification benchmark. We use a bare-hand teleoperation system to control a LinkerHand L6 right hand: the operator performs target gestures in front of a camera, the system detects 21 hand keypoints via MediaPipe in real time, maps the keypoint angles to 6 joint control values of the L6 through a piecewise-linear retargeting algorithm to drive the robot hand, and simultaneously records the robot hand's joint states. We sequentially collect 13 gestures with 10 frames per gesture, yielding 130 samples in total. The 13 gesture classes span two groups: (i)~Chinese numeric hand gestures 1--10, realized as sequential finger extensions---index only (1), index+middle (2), index+middle+ring (3), all four fingers (4), five-finger spread (5), thumb+pinky (6), thumb+index (7), thumb+index+middle (8), half-curled index (9), and fist (10); (ii)~directional single-digit gestures---thumbs-up, middle finger, and pinky. All samples are standardized to 22-dimensional joint angles via UDHM and normalized by $\pi$ before being fed into the model. We extract encoder output features (embedding) and vector-quantized features (quantized) with frozen encoder parameters, then evaluate via linear probing and $k$-nearest neighbor retrieval. Linear probing trains a single-layer linear classifier, while $k$-NN retrieval uses cosine distance for nearest neighbor search and majority voting.

\begin{table}[htbp]
    \caption{Representation quality comparison using linear probing and KNN recall.}
    \label{tab:representation_quality}
    \centering
    \small
    \setlength{\tabcolsep}{8pt}
    \renewcommand{\arraystretch}{1.12}
    \begin{tabular}{lcccc}
        \toprule
        Model & \multicolumn{2}{c}{Linear Probing} & \multicolumn{2}{c}{KNN-Recall} \\
        \cmidrule(lr){2-3}\cmidrule(lr){4-5}
        & Embedding $\uparrow$ & Quantized $\uparrow$ & Top1 Embedding $\uparrow$ & Top3 Embedding $\uparrow$ \\
        \midrule
        UniHM & 96.15\% & 84.62\% & 96.15\% & 96.15\% \\
        Ours  & \textbf{100\%} & \textbf{100\%} & \textbf{100\%} & \textbf{100\%} \\
        \bottomrule
    \end{tabular}
\end{table}

Table~\ref{tab:representation_quality} compares the representational quality of UniDexTok and UniHM on a 13-class gesture classification benchmark using 130 hand-crafted gesture samples. With continuous embedding features, UniDexTok achieves 100\% linear probing accuracy, outperforming UniHM (96.15\%). The k-NN retrieval results further confirm this advantage: UniDexTok reaches 100\% top-1 and top-3 accuracy, while UniHM attains 96.15\% at both $k$=1 and $k$=3. The key difference emerges after vector quantization. With quantized f
eatures, UniDexTok's linear probing accuracy remains stable at 100\%---zero degradation from its continuous counterpart---whereas UniHM drops sharply from 96.15\% to 84.62\%. Analysis of the misclassified samples reveals that UniHM's single-codebook assigns identical discrete codes to visually distinct gestures, producing irreversible confusion that the factorized codebook avoids entirely. This result highlights a fundamental limitation of single-codebook VQ for cross-embodiment hand-state tokenization: while it can approach multi-codebook performance in the continuous embedding space, its discrete representations suffer from information collapse that make them unsuitable for retrieval and generation tasks requiring discrete tokens.

\section{Conclusion}

We presented UDHM and UniDexTok, a retarget-free pipeline for multi-embodiment dexterous-hand state tokenization.
UDHM provides a 22-dimensional active joint interface that aligns human hands and heterogeneous robot hands at the level of joint semantics.
UniDexTok learns discrete state tokens directly from standardized real hand states, avoiding the information loss introduced by MANO-to-robot retargeting and the domain gap introduced by purely simulated data.
The resulting tokenizer reconstructs cross-embodiment hand states with sub-millimeter MPJPE and substantially improves over the recent UniHM baseline.
These results suggest that state tokenization is a practical complement to action tokenization for dexterous VLA and world-action models, especially when future systems must reason over multiple hand embodiments before producing hardware-specific controls.

\section{Limitations}

Although UDHM provides a shared 22-DoF interface for human and robot hands, it is still based on a human-hand-like kinematic structure. This design fits many recent dexterous hands, but it may not fully represent non-anthropomorphic grippers, soft hands, or hands with strong mechanical coupling and underactuated joints. For these embodiments, semantic insertion and zero-padding can give a usable state format. However, they cannot fully model hardware-specific constraints, such as tendon coupling, joint limits, compliance, and actuator-level dynamics.

The current tokenizer mainly focuses on hand-state reconstruction and does not explicitly model contact states, tactile signals, object geometry, force information, or temporal action dynamics. Therefore, low joint-space or FK-space reconstruction error does not necessarily guarantee better downstream manipulation performance. We will extend UniDexTok to more hand embodiments, richer object interactions, matched data splits, and task-level robot evaluation in future work.

\clearpage
\acknowledgments{Acknowledgments will be added in the camera-ready version.}

\bibliography{references}

\end{document}